\documentclass[a4paper]{article}

\usepackage{INTERSPEECH2022}
\usepackage{multirow}
\usepackage{xcolor}
\usepackage[a-1b]{pdfx} 

\title{What can Speech and Language Tell us About 
the \\ Working Alliance in Psychotherapy
}
\name{S.P. Bayerl$^1$, G. Roccabruna$^4$, S.A. Chowdhury$^3$,T. Ciulli$^2$, M. Danieli$^4$,\\K. Riedhammer$^1$, G. Riccardi$^4$}
\address{
  $^1$Technische Hochschule Nürnberg Georg Simon Ohm, Germany\\
  $^2$Digital Psychology, IDEGO, Italy,
  $^3$Qatar Computing Research Institute, Qatar,\\  
  $^4$Signals and Interactive System Lab, DISI, University of Trento, Italy}
\email{giuseppe.riccardi@unitn.it}

\begin{document}

\maketitle
\begin{abstract}
We are interested in the problem of conversational analysis and its application to the health domain. Cognitive Behavioral Therapy is a structured approach in psychotherapy, allowing the therapist to help the patient to identify and modify the malicious thoughts, behavior, or actions.
This cooperative effort can be evaluated using the Working Alliance Inventory Observer-rated Shortened -- a 12 items inventory covering task, goal, and relationship -- which has a relevant influence on therapeutic outcomes.
In this work, we investigate the relation between this alliance inventory and the spoken conversations (sessions) between the patient and the psychotherapist.
We have delivered eight weeks of e-therapy, collected their audio and video call sessions, and manually transcribed them. The spoken conversations have been annotated and evaluated with WAI ratings by professional therapists.
We have investigated speech and language features and their association with WAI items. The feature types include turn dynamics, lexical entrainment, and conversational descriptors extracted from the speech and language signals. Our findings provide strong evidence that a subset of these features are strong indicators of working alliance. To the best of our knowledge, this is the first and a novel study to exploit speech and language for characterising working alliance.


\end{abstract}
\noindent\textbf{Index Terms}: conversational analysis, working alliance, psychotherapy
\vspace{-0.2cm}
\section{Introduction}

Research interest in automatic behavioral analysis has been increasing in the past few years\cite{narayanan2013behavioral}. It has been proved that carefully designed, AI-enabled, Personal Healthcare Agents (PHA) \cite{riccardi2014towards} may improve many aspects of the treatment processes, including citizens’ access to healthcare services and improvement of patients’ adherence to medical prescriptions and wellbeing recommendations. 

To reach those benefits, it is crucial that the design and implementation of PHA may rely on the understanding of the human therapeutic process, collaborative design, and on the availability of high-quality resources like annotated corpora of patient-therapist verbal interactions. A critical factor for the success of the PHA is the patients’ engagement during the care process. In other words, like in human-human patient-therapist interactions, the establishment of a satisfactory therapeutic relationship. The different facets of such relationships have been investigated by psychologists and medical doctors in depth in the recent past \cite{pringle2015improving,pihlaja2018therapeutic}. The concept of therapeutic alliance has emerged as a candidate for operationally describing the multiple intertwined factors that characterize compliance with medical prescriptions, user satisfaction, and a high chance of reaching benefits when setting and pursuing personal health goals. While this concept is general and applicable to all domains of healthcare, it is particularly crucial in mental healthcare because of the relevance of the bond that may be established between the therapists and their patients. 

In this paper, we analyse patient-therapist spoken conversations collected over an eight-week Cognitive Behavioral Therapy (CBT) intervention in a real setting. The study aims at investigating the correlations between the conversation features and the patient-therapist working alliance. We describe the Working Alliance Inventory (WAI), the annotation process, and the inter-annotator agreement (IAA). The IAA results are very satisfactory and aligned with related studies using Cronbach’s Alpha as IAA metric \cite{herrero2020working}. 
We have automatically extracted turn-taking, with other basic speech (e.g. duration and token), and language features from the conversations and correlated with WAI ratings. 
The turn-taking feature correlations show that equality in the engagement and acknowledgement utterances are relevant indexes of a good working alliance. The important role of the engagement is also observed in the most correlated turn-level features, which represent the speech activity at the segmentation and conversation levels. Furthermore, we have observed the same results on the discourse side, by measuring patient-therapist engagement through lexical entrainment for different word groupings and dialogue acts of these segments. Our analyses provides a strong evidence that the working alliance is manifested in speech and language. This result could provide support to further research in automatic behavioural analysis. 

\section{Concepts underlying therapeutic relationship}
\subsection{Working Alliance}
The research on the evaluation of psychotherapy outcomes identified three key concepts that are frequently associated with good therapeutic results, measured in terms of satisfaction, acceptability, and symptom remission. The three concepts are proper information, cooperation, and commitment to achieve the objectives of treatment. While the concept of proper information pertains to the domain of ethics, cooperation and commitment to achieve treatment goals have been related to the construct of working (or therapeutic) alliance.

According to many authors, the working alliance are one of the strongest predictors of the successful outcome of a psychotherapeutic path \cite{horvath1991relation,martin2000relation}. In \cite{thompson2012effect}, the study shows the association between Alliance and Adherence, defined as the extent to which the patient’s behavior coincides with the medical or health advice.  

The concept of working alliance has developed over time, and through the various approaches that characterize the psychological pathways. The authors in \cite{horvath1993role} explain how this concept has been discussed in the perspective of the different theories, from its origin in the psychodynamic approach, up to the contemporary view as a pantheoretical concept. Following a pantheoretical way, the study \cite{bordin1979generalizability} identified the three underlying factors (related components). The three underlying factors of working alliance are the mutual understanding and trust that therapist and client may establish (bond), their willingness of setting a shared goal (goal), and the purposive collaboration between them (task). 

Due to the importance of this construct in the determination of better results, over the years many researchers proposed questionnaires to measure the working alliance, for example, recently the authors in \cite{herrero2020working} studied the psychometric properties of this inventory for its application to self-assessment in online psychotherapy interventions.

\subsection{Working Alliance Inventory}
The Working Alliance Inventory (WAI henceforth, \cite{horvath1989development}) is a questionnaire of items arranged on a Likert scale and is one of the most used to assess the alliance \cite{martin2000relation} according to the integrative model by Bordin \cite{bordin1979generalizability}. The three specific factors (bond, goal, and task) mentioned above were examined by a specific confirmatory factor analysis  \cite{tracey1989factor}, which allowed defining the 12-item short form of the inventory. The questionnaire has been validated both for self-assessment and for being administered by an observer \cite{tichenor1989comparison}, and both versions have been validated in several languages. In this study, we administered the Italian form of the WAI Observer-rated Shortened (WAI-O-S) version \cite{tichenor1989comparison} as reported in \cite{lingiardi2002alleanza}. 
\renewcommand{\arraystretch}{1.6}
\begin{table}[h!]
    \centering
    \begin{tabular}{c|p{0.75\linewidth}}
        \textbf{Factors} & \textbf{Items}\\
        \hline
        \multirow{4}{*}{\textbf{GOAL}}&\textbf{WAI 4:} There are doubts or a lack of understanding about what participants are trying to accomplish in therapy.\\
        \cline{2-2}
        &\textbf{WAI 6:} The client and therapist are working on mutually agreed upon goals.\\
        \cline{2-2}
        &\textbf{WAI 10:} The client and therapist have different ideas about what the client’s real problems are.\\
        \cline{2-2}
        &\textbf{WAI 11:} The client and therapist have established a good understanding of the changes that would be good for the client.\\
        \hline\hline
        \multirow{4}{*}{\textbf{TASK}}&\textbf{WAI 1:} There is agreement about the steps taken to help improve the client’s situation.\\
         \cline{2-2}
        &\textbf{WAI 2:} There is agreement about the usefulness of the current activity in therapy (i.e., the client is seeing new ways to look at his/her problem).\\
        \cline{2-2}
        &\textbf{WAI 8:} There is agreement on what is important for the client to work on.\\
        \cline{2-2}
        &\textbf{WAI 12:} The client believes that the way they are working with his/her problem is correct.\\
        \hline\hline
        \multirow{4}{*}{\textbf{BOND}}&\textbf{WAI 3:} There is mutual liking between the client and therapist.\\
         \cline{2-2}
        &\textbf{WAI 5:} The client feels confident in the therapist’s ability to help the client.\\
        \cline{2-2}
        &\textbf{WAI 7:} The client feels that the therapist appreciates him/her as a person.\\
        \cline{2-2}
        &\textbf{WAI 9:} There is mutual trust between the client and therapist.\\
        \hline
    \end{tabular}
    \caption{WAI items and their categorization in factors \cite{tracey1989factor}.}
    \label{tab:wai_items}
    \vspace{-1.0cm}
\end{table}
\vspace{-0.2cm}
\section{Corpus}
\subsection{Description}
The corpus is a collection\footnote{This data collection has been approved by the Ethical Committee of the University of Trento.} of 23 Italian dyadic conversations during the sessions of Cognitive Behavioral Therapy (CBT). 
The intervention can last out from 5 to 20 sessions, each session of $\approx 1$ hours, in which the patient works with the therapist on getting a better understanding of his/her issues. CBT therapy can reduce the impact on the patient's life of many symptoms, such as phobia, depression, anxiety, and bipolar disorders. In this corpus,  three patients attended, via voice and video call, 8 therapy sessions. However, for one of the patients, the first session is missing.  

The patient-therapist dialogues were manually transcribed. In all, the corpus is composed of 5.9K turns, the dictionary size is 8.1K words and the average turn length is 12 tokens. The dialogues were segmented into turns using the silent pauses between utterances, thus multiple consecutive turns can belong to the same speaker.

\subsection{Annotation task}
The annotation task consisted of listening to the spoken conversation of a session and at the end of it, the annotators were asked to fill the WAI-O-S questionnaire by replying to the 12 questions listed in Table \ref{tab:wai_items},  by choosing a single option selection on a Likert scale from 1 (never) to 7 (ever).
Two CBT experts performed the annotation of these therapeutic conversations.
Based on those annotations, we computed the inter-annotator agreement (IAA) and the correlations to dialogue features.

The IAA has been computed with the Cronbach’s alpha \cite{cronbach1951coefficient} metric; according to its interpretation table \cite{gliem2003calculating}, we report: 15 questionnaires achieve excellent internal consistency ($\alpha >  0.9$), 4 good consistency ($0.9 > \alpha \geq 0.8$), 1 acceptable ($0.8 > \alpha \geq 0.7$), 1 questionable ($0.7 > \alpha \geq 0.6$) , 1 poor ($0.6 > \alpha\geq 0.5$) and 1 unacceptable ($\alpha \geq 0.5$). These values fall in line with other studies \cite{herrero2020working}. 

\section{Conversational Features}
Speech-based cues offer an important means of measuring cooperative efforts between the therapist and the patient. Moreover, these verbal cues hold the key to describe the turn-taking (interaction) dynamics between the interlocutors. 
Turn-taking is a remarkable phenomenon that represent the essence of human communication \cite{julia2011pragmatic}. Over decades these intriguing cues are utilised to predict the outcome of conversations and dominance of speakers among others \cite{chowdhury2016predicting,julia2011pragmatic,chowdhury2019automatic}. 
Motivated by its success to capture the flow of conversation,
we exploit these features for understanding the working alliance between the therapist and the patient. For this study, we investigate the phenomena in two stages using: \textit{(i)} turn-taking measure incorporating conversational discourse and turn dynamics; \textit{(ii)} turn-level measure where we studied windowed or individual segments in the conversation.

\subsection{Methods and Features}
\paragraph*{Turn-taking Features} We extracted the turn-taking features by aligning outputs from speaker diarization sub-modules \cite{Bredin2020}. For the task, we utilised voice activity, overlap and speaker change detection using the raw dyadic audio. We aligned these automatic turn segments (short spurts/segments separated by at least 500ms silence) with the manual segmentation\footnote{Segmentation used for manual transcription.} and then assigned manually annotated speaker role. 

Using these segments, we extracted global features to understand conversation structure and participation. The features includes: 
\begin{enumerate} \setlength\itemsep{-0.0em}
    \item \textbf{Participation equality} \cite{lai2013modelling, chowdhury2016predicting} -- measuring the difference in the proportion of turn duration hold by each speaker.  
We defined, $P_{eq}=1-(\frac{\sum _{i}^{N} (T_i-T)^{2}/T}{E})$
where $T$ is the average speech duration of the speakers. $T_i$ is the total speech duration for each speaker. $E$ and $N$ represent the maximum possible duration a speaker can hold and number of speakers. The value of $P_{eq}$ closer to 1 indicate greater equality between speakers.
\item \textbf{Turn-level Freedom}, \cite{lai2013modelling, chowdhury2016predicting} -- measuring the predictability of turn-taking structure, defined by $F_{cond}=1-\frac{H_{max}(Y|X)-H(Y|X)}{H_{max}(Y|X)}$, where we calculated $H(Y|X)$, the conditional entropy of speaker $Y$ being the next speaker after $X$ begins the turn, $H_{max}(Y|X)$ being the maximal possible value for this. $W=\{therapist, patient\}$, $X\in W$, $Y\in W$ and $X \neq Y$. The value of $F_{cond}$ is in between 0 and 1, where 0 represents a deterministic turn-taking (i.e. only speaker Y follows X strictly).
\item \textbf{Overlapping turns} \cite{chowdhury2015role, chowdhury2019automatic} 
-- measuring the number and percentage of overlapping turns in the conversation.
\vspace{-0.2cm}
\end{enumerate}

\paragraph*{Turn-level Features} As the session unfolds, the turn properties of the speakers changes. We have observed large differences in positional features such as the mean turn duration of therapists and patients which ranged from 12.6 to 22.8 seconds for the therapist and 8.0 to 14.6 seconds for the patient. 

Motivated by this, we extracted per speaker turn duration, token count, speech rates and pause-based (between and with-in the speakers) features for different parts (beginning/center/end), as well as the total percentage of speaking time for the interlocutors during the whole conversation like.
For the study, we defined the beginning and end as the first and last 15\% of the conversation (duration), the remaining part as center/middle position of the conversation.

\begin{table}[]
\centering
\begin{tabular}{l|c|c}
\multicolumn{1}{c|}{\textbf{Features}} & \textbf{WAI} (Factors)& \textbf{Corr} (Slope) \\ \hline\hline
\multirow{5}{*}{\textit{Participation Equality}} & WAI 1 (Task) & 0.47 (2.59) \\ \cline{2-3} 
 & WAI 2 (Task) & 0.50 (4.16) \\ \cline{2-3} 
 & WAI 5 (Bond) & 0.42 (3.49) \\ \cline{2-3} 
 & WAI 11 (Goal) & 0.44 (3.36) \\ \cline{2-3} 
 & WAI 12 (Task) & 0.53 (4.72) \\ \hline\hline
\multirow{4}{*}{\begin{tabular}[c]{@{}l@{}}\textit{Turn Freedom:} \\ \textit{Therapist$^*$}\end{tabular}} & WAI 1 (Task) & 0.43 (1.49) \\ \cline{2-3} 
 & WAI 2 (Task) & 0.54 (2.81) \\ \cline{2-3} 
 & WAI 8 (Task) & 0.53 (1.92) \\ \cline{2-3} 
 & WAI 12 (Task) & 0.51 (2.87)\\ \hline\hline
\multirow{2}{*}{\begin{tabular}[c]{@{}l@{}}\textit{Minimal Speech Rate}:\\ \textit{Therapist (tokens per sec)}\end{tabular}} & WAI 9 (Bond) & 0.69 (0.25) \\ \cline{2-3} 
 & WAI 8 (Task) & 0.60 (0.28) \\ \hline\hline
 \multirow{3}{*}{\begin{tabular}[c]{@{}l@{}}\textit{Number of} \\ \textit{Overlapping Turns}\end{tabular}} & WAI 2 (Task) & 0.42 (0.03)\\ \cline{2-3} 
 & WAI 5 (Bond) & 0.42 (0.02) \\ \cline{2-3} 
 & WAI 12 (Task) & 0.47 (0.03)\\ \hline
\end{tabular}%
\caption{Reported significant Pearson correlation of turn-taking features (\textit{p-value} $< 0.05$ ) with WAI ratings and relative regression slopes (Slope).}
\label{tab:turn-taking-feats}
\vspace{-0.9cm}
\end{table}
\vspace{-0.2cm}
\subsection{Results}
The estimated Pearson correlation (with \textit{p-value} $<0.05$), along with the regression slope, between the WAI-O-S ratings and the global turn-taking measures are reported in Table \ref{tab:turn-taking-feats}. The results indicate that an equal engagement ($P_{eq}$) from both interlocutors can lead to better working alliance for Task (\textbf{WAI 1}, \textbf{WAI 2} and \textbf{WAI 12}), Goal (\textbf{WAI 11}) and Bond (\textbf{WAI 5}).

The result also pinpoints the more unpredictable turn-taking strategy the therapist used (for e.g., giving short feedback signals, showing agreement between patient's turns, or giving patient more speaking time) often leads to better alliance scores for all the Task factors. 
Furthermore, we observed a slight positive effect of increase number of short overlapping spurt with Task (\textbf{WAI 2, WAI 12}) and Bond (\textbf{WAI 5}). This suggests that short feedback utterances are important for modelling a healthy alliance between the speakers.  

Moreover, we observed a positive correlation between the minimum observed rate of speech of the therapist and the score for Task (\textbf{WAI 8}) and Bond (\textbf{WAI 9}). We hypothesize that the more mutual trust observed between patient and therapist, the therapist is more comfortable to talk faster. This also shows that the therapist is well-aware of the patient's agreement towards the treatment plan.

In addition to the global features, we also observed correlation with other positional turn information.  We noticed, the median duration of the patient speaking in any observed time window is positively
correlated ($\rho=0.64$) with the belief that the patient and therapist are working on his/her problems correctly (\textbf{WAI 12}). We hypothesize that the more regular the client engages in the conversation with the therapist, the more he/she is motivated to be working on his/her problems correctly. 

We also observe a significant negative correlation ($\rho=-0.55$) between the standard deviation of the therapists' turn duration and the rating for \textbf{WAI 11}. The more variation there is in therapists' turn duration, the less common understanding there is between patient and therapist on the changes that would benefit the patient.  This could mean, long stretches of the therapist talking to the patient (schooling/telling) instead of talking with the patient, followed by a time that has more of a conversation characteristic. This hypothesis is also backed by the correlation coefficient seen between total percentage of turn duration belonging to patient ($\rho=0.58$ for \textbf{WAI 12} and $\rho=0.43$ for \textbf{WAI 11}). \footnote{A Negative correlation is seen with therapist's total percentage of turn duration.} 

There also is a positive correlation between the total duration the patient talks in the center part of the conversation  (from 15\% to 85\% of the total duration, and the annotated rating for \textbf{WAI 5}. Meaning the more the client talks (engages in the conversation) during the middle part of the therapy session the more confidence he or she has that the therapist can help him or her.

\section{ Lexical Entrainment}
The lexical entrainment is the phenomena in which a speaker conveys information reusing terms spoken by the other interlocutor. Furthermore, \cite{hirschberg2008high} defined a metric to estimate the entrainment, which effectiveness was tested on predicting the perceived naturalness of a dialogue. 

\noindent The metric is defined as:
\begin{equation}\label{eq:entrain}
    ENTR(c) = -\frac{\sum_{w \in c}|count_{S_1}(w) - count_{S_2}(w)|} {\sum_{w \in c}|count_{S_1}(w) + count_{S_2}(w)|}
    \vspace{-0.1cm}
\end{equation}

\noindent where $c$ is a target word class and $count_{S_i}$ is the frequency of the word $w$  used by the speakers $S_1$ and $S_2$. The word frequencies are computed over the set of dialogues selected for the estimation of the entrainment. We have used the entrainment metric to analyze possible correlations with the normalized WAI ratings and entrainment rating. We hypothesize that WAI ratings, which assess the working alliance between patient and therapist, should have a relation with entrainment metrics measuring user engagement. 

\subsection{Methods and Results}
In previous research \cite{hirschberg2008high}, the word classes built to compute the entrainment were based on word frequencies, for instance, selecting the top 25 or 100 most frequent words of the whole corpus. In Table \ref{tab:top_25}, we report the Pearson correlations between entrainment and WAI ratings for the top-25 most frequent words across the whole corpus (23 conversations). These words are reported in the first row of Table \ref{tab:top_25}. 
We have further extended this analysis to all the words in the dictionary. We compute entrainment scores for each patient’s conversation. For each patient, we sort by frequency the words occurring in his/her conversations.  We have then processed the sorted word list with a sliding window of 100 words and an overlap of 85 words. This process has generated 66 word-bins per patient. For each bin, we compute the entrainment scores (eq. \ref{eq:entrain}) for each patient’s conversation. We correlate the entrainment scores, using Pearson correlation, with the 23 WAI ratings of a particular question. Due to the overall good inter-annotator agreement, we averaged the WAI ratings of the two annotators. Table \ref{tab:ent_corr} reports the highest correlation values amongst all word bins and WAI items. We have analyzed the part-of-speech of these words, using Spacy part-of-speech tagger, and have found 75\% of the words are content words, i.e. they fall in one of these categories nouns, verbs, adjectives, adverbs, and main verbs. In the first column of Table \ref{tab:ent_corr}, we show representative words in the bin for the corresponding WAI question. We have inspected these words and the context they appear. We manually categorize these words in terms of their discourse function according to the Dialogue Act standard ISO-24617-2 \cite{bunt2012iso}. From this, we observe that \textit{inform}, \textit{agreement} and \textit{offer} discourse functions correlate well with Task \textbf{(WAI 1, WAI 12, WAI 2}). While \textit{feedback} and \textit{request} discourse functions correlate with Goal (\textbf{WAI 11}) and Bond (\textbf{WAI 5}), respectively.  
\begin{table}[h!]

    \centering
\resizebox{\linewidth}{!}{
\begin{tabular}{c|lllll}
\textbf{\begin{tabular}[c]{@{}c@{}}Top-25\\  words\end{tabular}} & \multicolumn{5}{l}{\begin{tabular}[c]{@{}l@{}}“che”, “è”, “di”, “non”, “un”, “e”, “in”, “a”,  \\“si”,“la”, “il”, “mi”, “per”, “anche”, “una”,\\“l'”, “perché”,  “quindi”, “sono”,  “questo”,\\ “se”, “ho”, “ci”, “le”, “ha”\end{tabular}} \\ \hline
\begin{tabular}[c]{@{}c@{}}\textbf{WAI} \\ (Factors)\end{tabular} & \multicolumn{1}{l|}{\begin{tabular}[c]{@{}l@{}}WAI 12 \\ (Task)\end{tabular}} & \multicolumn{1}{l|}{\begin{tabular}[c]{@{}l@{}}WAI 11 \\ (Goal)\end{tabular}} & \multicolumn{1}{l|}{\begin{tabular}[c]{@{}l@{}}WAI 5 \\ (Bond)\end{tabular}} & \multicolumn{1}{l|}{\begin{tabular}[c]{@{}l@{}}WAI 2 \\ (Task)\end{tabular}} & \begin{tabular}[c]{@{}l@{}}WAI 1 \\ (Task)\end{tabular} \\ \hline
\textbf{Corr} & \multicolumn{1}{l|}{0.52}& \multicolumn{1}{l|}{0.46} & \multicolumn{1}{l|}{0.46} & \multicolumn{1}{l|}{0.45} & \multicolumn{1}{l}{0.44} \\
(Slope) & \multicolumn{1}{l|}{(1.17)} & \multicolumn{1}{l|}{(1.32)} & \multicolumn{1}{l|}{(1.45)} & \multicolumn{1}{l|}{(1.41)} & \multicolumn{1}{l}{(0.91)} \\ \hline

\end{tabular}}%
    \caption{This table reports the top-5 Pearson correlations (p-value \textless{} 0.05), along with regression slopes (Slope), between the entrainment ratings and WAI ratings. The word group used to compute the entrainment ratings includes the 25 most frequent words, listed in ``Top-25 words'' row,  across the entire corpus.}
    \label{tab:top_25}
    \vspace{-0.8cm}
\end{table}
\begin{table}[h!]
    \centering
    \begin{tabular}{p{0.6\linewidth}|p{0.145\linewidth}|p{0.10\linewidth}}
    \textbf{Dialogue Act}& \textbf{WAI} (Factors) &\textbf{Corr} (Slope)\\ \hline
         \textbf{Inform}: ``dicevi'' (in therapeutic conversations it is used for promoting self-attention)& WAI 1 (Task) & 0.64 (2.59)\\
         \hline
         \textbf{Agreement}: ``infatti'', ``effettivamente'', ``assolutamente''& WAI 12 (Task) &0.64 (2.45)\\
         \hline
         \textbf{Offer}: ``proviamo'',  ``vediamo'', ``facciamo'', ``possiamo''& WAI 2 (Task) &0.63 (3.41)\\
         \hline
         \textbf{Feedback functions}: ``mah'', ``beh'', ``benissimo'', ``esattamente'' &WAI 11 (Goal) & 0.62 (3.43)\\
         \hline
         \textbf{Request}: ``pensa'', ``aspetta'', ``devi''& WAI 5 (Bond) & 0.61 (3.71)\\
         \hline
    \end{tabular}
    \caption{Top-5 Pearson correlations (p-value \textless{0.05}), along with regression slopes (Slope), between WAI ratings and entrainments ratings of word groupings. The groupings were computed by a sliding window of 100 words wide with an overlap of 85 along the corpus word list sorted by frequency. In the column,``Dialogue Act'' we select a sample of the most relevant words in that grouping and their associated dialogue act as per ISO-24617-2 standard \cite{bunt2012iso}.}
    \label{tab:ent_corr}
    \vspace{-0.9cm}
\end{table}

\vspace{-0.2cm}
\section{Conclusions}

We have presented the analysis of patient-psychotherapist conversations collected during an eight-week CBT intervention on three patients.
We described the WAI assessment scheme of the working alliance, and then detailed the annotation task of these spoken conversations.
Following, we analysed features extracted from speech and language cues to find correlations with the WAI ratings. We empirically showed that turn-taking structure in the session is evidently a predictor of WAI.
We observed that encouraging participation equality and more patient turn time is a strong feature for higher WAI ratings. 
From the language side, we noticed higher use of \textit{feedback}, along with \textit{inform}, \textit{request} discourse functions correlates significantly with higher WAI ratings. These results provide insights into the future work on automatically computing the working alliance in therapeutic sessions and extension to other health domains.

\vspace{-0.2cm}
\section{Acknowledgements}
The research leading to these results has received funding from H2020 Grant Agreement 826266: COADAPT.

\bibliographystyle{IEEEtran}

\bibliography{mybib}

\end{document}